\documentclass[letterpaper, 10pt,conference]{ieeeconf}  

\IEEEoverridecommandlockouts                         
\makeatletter
\let\NAT@parse\undefined
\makeatother

\usepackage{makecell} 
\usepackage[square, comma, numbers,sort]{natbib}
\usepackage{amsmath,amssymb,amsfonts}
\usepackage{algorithmic}
\usepackage{graphicx}
\usepackage{textcomp}
\usepackage{multirow}
\usepackage{bbding} 
\usepackage{pifont}
\usepackage{booktabs}
\usepackage{balance}
\usepackage{color}
\usepackage[table,dvipsnames]{xcolor}

\usepackage[pagebackref=true,breaklinks=true,colorlinks,bookmarks=false]{hyperref}

\hypersetup{
linkcolor=BrickRed
,citecolor=Green
,filecolor=Mulberry
,urlcolor=NavyBlue
,menucolor=BrickRed
,runcolor=Mulberry
,linkbordercolor=BrickRed
,citebordercolor=Green
,filebordercolor=Mulberry
,urlbordercolor=NavyBlue
,menubordercolor=BrickRed
,runbordercolor=Mulberry
}

\def\BibTeX{{\rm B\kern-.05em{\sc i\kern-.025em b}\kern-.08em
    T\kern-.1667em\lower.7ex\hbox{E}\kern-.125emX}}

\begin{document}
\author{Suozhi Huang$^{1,2,*}$, Juexiao Zhang$^{1,*}$, Yiming Li\textsuperscript{1,\ding{41}}, and Chen Feng\textsuperscript{1,\ding{41}}
\\
{\tt\small\url{https://coperception.github.io/ActFormer/}}
\thanks{* indicates equal contributions. Work done during Suozhi’s visit at NYU.}
\thanks{\ding{41} Corresponding authors. This work is supported by NSF Grant 2238968. This work was supported in part through the NYU IT High Performance Computing resources, services, and staff expertise.}
\thanks{$^{1}$Juexiao Zhang, Yiming Li, and Chen Feng are with New York University, New York, NY 11201, USA {\tt\small \{juexiao.zhang, yimingli, cfeng\}@nyu.edu}}
\thanks{$^{2}$Suozhi Huang is with Institute for Interdisciplinary Information Sciences (IIIS), Tsinghua University, Beijing 100084, China  {\tt\small huang-sz20@mails.tsinghua.edu.cn}}
}

\title{\LARGE \bf
    ActFormer: Scalable Collaborative Perception via Active Queries}

\maketitle
\thispagestyle{empty}
\pagestyle{empty}

\begin{abstract}
    Collaborative perception leverages rich visual observations from multiple robots to extend a single robot's perception ability beyond its field of view.
    Many prior works receive messages broadcast from all collaborators, leading to a scalability challenge when dealing with a large number of robots and sensors.
    In this work, we aim to address \textit{scalable camera-based collaborative perception} with a Transformer-based architecture. Our key idea is to enable a single robot to intelligently discern the relevance of the collaborators and their associated cameras according to a learned spatial prior. This proactive understanding of the visual features' relevance does not require the transmission of the features themselves, enhancing both communication and computation efficiency. Specifically, we present ActFormer, a Transformer that learns bird's eye view (BEV) representations by using predefined BEV queries to interact with multi-robot multi-camera inputs. Each BEV query can actively select relevant cameras for information aggregation based on pose information, instead of interacting with all cameras indiscriminately. Experiments on the V2X-Sim dataset demonstrate that ActFormer improves the detection performance from 29.89\% to 45.15\% in terms of AP@0.7 with about 50\% fewer queries, showcasing the effectiveness of ActFormer in multi-agent collaborative 3D object detection.
\end{abstract}

\section{Introduction}
Collaborative perception, such as collaborative object detection~\cite{li2021learning} and semantic segmentation~\cite{xu2022cobevt}, empowers autonomous robots to share their perceptual insights, fostering a comprehensive understanding of their surrounding environments. It addresses challenges such as occlusions and sparse sensory information over long distances, which often impede individual perception. Nevertheless, scalability poses a notable challenge to current learning-based collaborative perception methods. This challenge mainly stems from the passive nature of existing approaches, which incorporate all accessible sensor data at some point, rather than \textit{actively} requesting only the essential information before initiating any form of communication with collaborators.

\begin{figure}[t]
    \centering
    \includegraphics[width=\linewidth]{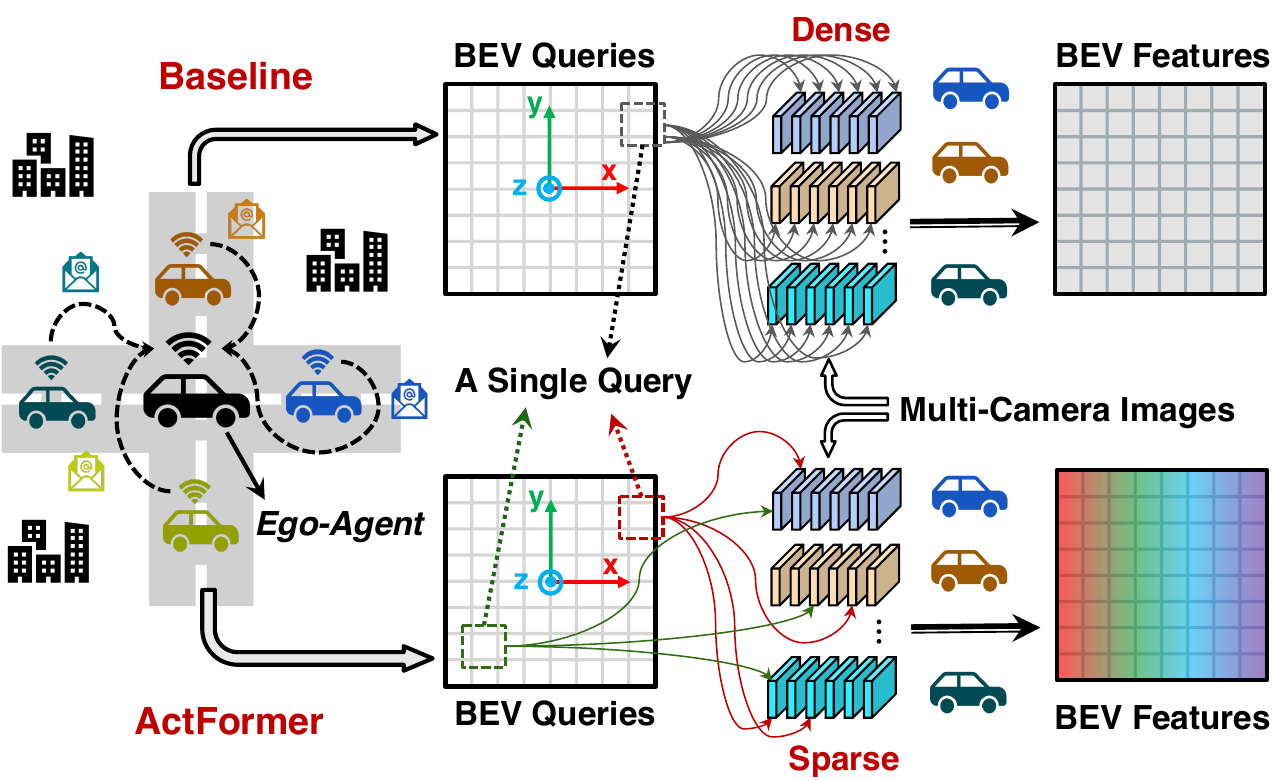}
    \caption{\textbf{Baseline v.s. ActFormer.} Baseline densely attends BEV queries to every 2D image feature. ActFormer actively selects queries based on spatial information and achieves more scalable collaborative perception.
    }
    \label{fig:teaser}
    \vspace{-5mm}
\end{figure}

Certain prior approaches \cite{liu2020when2com,liu2020who2com,hu2022where2comm} involve selective data usage to minimize communication overhead. However, the decision-making process for selecting sensor data still relies on transferred feature representations of the sensor measurements from other collaborators. As a result, these approaches are not truly \textit{active} and struggle to scale up effectively when dealing with a large number of sensors or robots. Hence, this work aims to address the problem of \textit{scalable collaborative 3D perception based on 2D camera input}. We propose to enable the ego-robot to \textit{actively and intelligently request and use information}, as opposed to \textit{passive and indiscriminate utilization of all available camera images}.

Inspired by the recent advances of Transformer for camera-based single-robot perception~\cite{wang2022detr3d, li2022bevformer}, we design our method based on this query-based framework that leverages a learnable grid-shaped bird's eye view (BEV) queries corresponding to the target 3D space, to interact with the 2D camera features for 3D representation learning. Yet existing query-based methods like BEVFormer~\cite{li2022bevformer} are not designed to handle scalable camera-based collaborative perception. Therefore, each BEV query is supposed to interact with multi-agent multi-camera input in a dense manner, severely limiting the scalability and efficiency of the collaborative perception system, as shown in Fig.~\ref{fig:teaser} (top branch).

In practice, querying all available sensory streams densely is highly redundant and inefficient. Our key idea is to optimize the query graph by enabling each BEV query to identify the most relevant 2D camera features based on the poses of robots and their cameras. To implement this idea effectively, we employ a learnable active selection module. This module takes input in the form of the ego robot's BEV features and the poses of collaborators, generating an interest score for each BEV query with respect to the available cameras. Only the high-score queries will be involved in the computation during the collaboration. Our method, termed \textit{ActFormer}, creates a sparse query graph for extracting a robust BEV representation from multi-robot multi-camera input, as illustrated in Fig.~\ref{fig:teaser} (bottom branch), largely boosting both efficiency and scalability for collaborative perception. We test ActFormer in the task of collaborative 3D object detection from 2D images and conduct comprehensive experiments on the widely-used V2X-Sim~\cite{li2022v2x} dataset. Quantitative and qualitative results show that ActFormer not only boosts the detection performance by a large margin compared to the baseline method ($45.88\% \rightarrow 53.33\%$ for AP@0.5;
$29.89\% \rightarrow 45.15\%$ for AP@0.7), but also reduces the computational cost with 50\% fewer queries. In summary, our major contributions are summarized as follows:
\begin{itemize}
    \item We conceptualize a scalable and efficient collaborative perception framework that can actively and intelligently identify the most relevant sensory measurements based on spatial knowledge, without relying on the sensory measurements themselves.
    \item We ground the concept of the scalable collaborative perception with a Transformer, \textit{i.e.}, \textit{ActFormer}, which uses a group of 3D-to-2D BEV queries to actively and efficiently aggregate the features from multi-robot multi-camera input, only relying on pose information.
    \item We conduct comprehensive experiments in the task of collaborative object detection to verify the effectiveness and efficiency of our ActFormer.
\end{itemize}

\begin{figure*}[t]
    \centering
    \includegraphics[width=\linewidth]{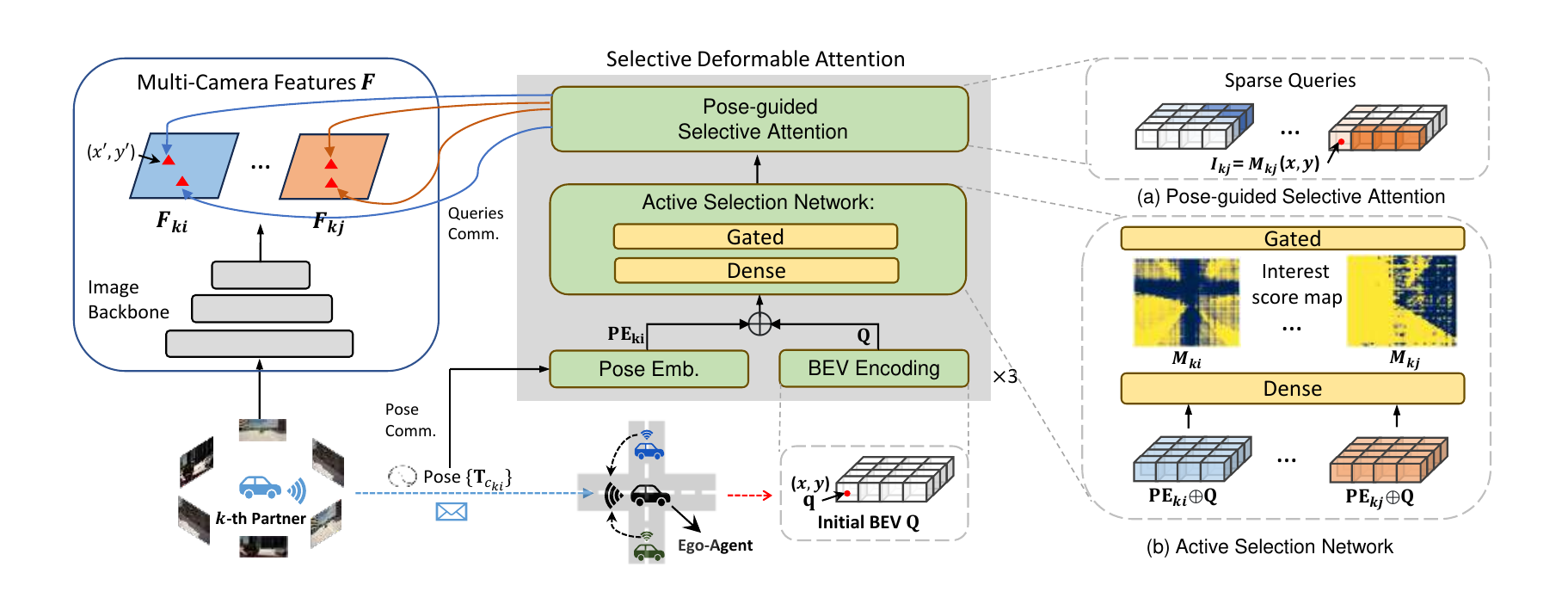}
    \vspace{-5mm}
    \caption{\textbf{Method overview.} After partners broadcast their pose information to the ego car, our approach leverages selective deformable attention to obtain active sparse queries for images. Selective deformable attention consists of two crucial components: (a) \textbf{Pose-guided Selective Attention}, which efficiently focuses on multi-agent image features using active queries, enabling us to emphasize regions of interest; and (b) \textbf{Active Selection Network}, which concatenates pose embeddings with BEV queries and produces an interest score map. Subsequently, this interest score map is multiplied by the BEV query using a gated network to obtain the active query. This process aims to enhance collaboration efficiency and generate active sparse queries.
    }

    \label{fig:main}
\end{figure*}

\section{Related Works}

\subsection{Collaborative perception}
Collaborative perception has been shown to enhance the adaptability, robustness and efficiency of individual autonomous vehicle systems~\cite{hu2022where2comm, lei2022latency, su2023uncertainty, 10377144, Hu_2023_CVPR,xu2022cobevt,su2024collaborative,li2023collaborative}. Many of them deploy feature-level collaborative perception techniques, where the intermediate features produced by the deep neural networks are shared among a team of robots, be it as a swarm of drones~\cite{liu2020when2com, zhou2022multi} or a fleet of autonomous vehicles~\cite{wang2020v2vnet, li2023multi, chen2019cooper}. Due to the high dimensionality of the intermediate features, collaboration cost is a critical issue to be considered. When2com~\cite{liu2020when2com} applies an attention mechanism over all the potential collaborators to fuse their features. Who2com~\cite{liu2020who2com} deploys a multi-stage handshake mechanism where the ego agent determines who to connect with based on the first stage of information exchange. V2VNet~\cite{wang2020v2vnet} trains a graph neural network for information aggregating. DiscoNet~\cite{li2021learning} learns about message selection via knowledge distillation~\cite{hinton2015distilling} from an oracle model that is trained with the complete scene observation. STAR~\cite{li2023multi} learns to amortize communicating the intermediate representations over a few consecutive time steps with a spatio-temporal masked autoencoder~\cite{feichtenhofer2022masked}.

Where2comm~\cite{hu2022where2comm} shares with us a similar insight to use spatial information to guide collaboration, but its confidence map is based on LiDAR encodings and it still requires obtaining the feature representations of the sensor measurements from collaborator robots. Meanwhile, most methods focus on the input of 3D LiDAR point clouds instead of 2D camera images. CoCa3D~\cite{Hu_2023_CVPR} tackles camera-based detection with the help of depth estimation. However, they did not address the scalability and efficiency issues. In a word, it is still underexplored how to actively request and exploit the sensor information of neighboring robots for scalable collaborative 3D perception from 2D camera images, without relying on the sensor measurements themselves.

\subsection{Transformer-based camera-only 3d perception}
3D perception tasks have been central to autonomous driving research~\cite{geiger2012kitti,chen2016monocular, roddick2018orthographic, mousavian20173d, kehl2017ssd, ku2019monocular}.
Recently, Transformers~\cite{vaswani2017attention} are successfully adopted into camera-only 3D perception tasks, including but not limited to 3D object detection~\cite{wang2022detr3d}, semantic scene completion~\cite{li2023voxformer}, \textit{etc}. Many of these Transformer-based models achieve superior performance by modeling an intermediate BEV representation before producing the task-specific outputs.
\cite{philion2020lift, reading2021categorical, huang2022bevdet4d} estimate the depth distribution then  project the 2D image features to 3D space to get the BEV
voxel features. Some treat the grid-shaped BEV features as queries or proposals and apply attention mechanisms over them and input features~\cite{xie2021segformer,zhou2022crossview}. Their BEV features are directly associated with the task outputs such as the object bounding boxes or voxel labels.

However, when involved with BEV features, the multi-head attention mechanism in the original Transformer can induce expensive computational costs since the number of queries is likely to be large.
To overcome such computational bottleneck, deformable attention~\cite{xia2022vision} is proposed which changes the original attention head from attending every token to only some of the tokens by learnable offsets:
\begin{equation}
    \operatorname{DfmAttn}(q, p, F)=\sum_{m=1}^{N_{\text {head }}} \mathcal{W}_m \sum_{n=1}^{N_{\text {key }}} \mathcal{A}_{m n} \cdot \mathcal{W}_m^{\prime} F\left(p+\Delta p_{m n}\right)
    \label{eq: dfm}
\end{equation}
where $q, p, F$ represent the query, reference point, and visual features, respectively. $N_{\text {head }}$ denotes the number of attention heads and $N_{\text {key }}$ is the number of sampled keys per attention head. $\mathcal{W}_m \in \mathbb{R}^{C \times\left(C / H_{\text {head }}\right)}$ and $\mathcal{W}_m^{\prime} \in \mathbb{R}^{\left(C / H_{\text {head }}\right) \times C}$ are the learnable weights, where $C$ is the feature dimension. $A_{m n} \in[0,1]$ is the normalized attention weight $\sum_{n=1}^{N_{\text {key }}} A_{m n}=1$ . $\Delta p_{m n} \in \mathbb{R}^2$ are the predicted offsets to the reference point $p$. $F\left(p+\Delta p_{m n}\right)$ represents the feature at location $p+\Delta p_{m n}$ in an image, extracted by bilinear interpolation as in~\cite{dai2017deformable}.
It is adopted from~\cite{zhu2020deformable} and extended to BEV representation learning in the BEVFormer~\cite{li2022bevformer}, which is the state-of-the-art method in camera-only 3D object detection.

In BEVFormer~\cite{li2022bevformer}, 3D query points are projected from 3D BEV space to 2D image space. Denote the BEV queries set as grid-shape $Q \in \mathbb{R}^{H\times W\times C}$ then $q \in \mathbb{R}^{1\times C}$ represents the query at a spatial location $(x,y)\in \mathbb{R}^{H\times W}$, denoted as $q = Q(x,y)$. Then the spatial cross attention between the query $q$ and image features:
\begin{equation}
    \operatorname{SCA}\left(q, F\right)=\frac{1}{\left|\mathcal{R}_{\text {hit }}\right|} \sum_{i \in \mathcal{R}_{\text {hit }}} \sum_{j=1}^{N_{\text {ref }}} \operatorname{DfmAttn}\left(q, \mathcal{P}(q, i, j), F_i\right)
    \label{eq:sca}
\end{equation}
note that only the images whose field of view is hit by query $q$, denoted as the set $\mathcal{R}_{\text {hit }}$, are considered.
For each BEV query $q$, project function $\mathcal{P}(q, i, j)$ gets the $j$-th reference point on the $i$-th view image by transforming with matrix $\mathbf{T_i}$. And $N_{\text{ref}}$ indicates the number of reference points along a z-axis on each BEV query position.
Hence for each query $q$, a pillar of 3D reference points $(x,y,z_j)^{N_{\text{ref}}}_{j=1}$ are projected to different image views through the specific projection matrix of cameras, which can be
written as:
\begin{equation}
    p = \mathcal{P}_k(q, i, j) = \mathbf{T_i} (x,y,z_j)^\intercal = \left(x^{\prime}_{i j}, y^{\prime}_{i j}\right)  ~\forall q,i,j
\end{equation}
$x^{\prime}_{i j}, y^{\prime}_{i j}$ are in the image space and $p$ is the 2D reference point as in Eq.~\ref{eq: dfm}.
Our ActFormer is built based on deformable attention. We extend the above SCA with an active selection mechanism that further selects queries based on the current vehicles' poses and enables scalable and efficient multi-agent collaboration.

\section{Method
 }

Since the task focuses on autonomous driving, we refer to the collaborating robots as \textit{vehicles} for clarity. Specifically, during the collaboration, any vehicle, referred to as the \textit{ego} vehicle when considering its perception, can actively select information from other vehicles, referred to as the \textit{partners}, to attend to, based on its current Bird's Eye View (BEV) representation and the relative poses of the others. Then the ego vehicle collaborates on active BEV queries with the corresponding partners' features and updates its own BEV representation accordingly. Finally, it decodes the detection output based on the updated BEV representation.

\subsection{Motivation} 
Imagine a scenario where an ego vehicle encounters a complex urban intersection with multiple partner vehicles in the same scene. Each vehicle is equipped with multiple cameras placed at various poses. Collaborative perception enables these vehicles to share their observations and achieve a higher level of perception than they could individually.
Our motivation stems from the idea that how vehicles collaboratively perceive should be closely related to their relative poses. Different camera poses result in varying viewpoints, each capturing unique information. However, conventional collaborative methods often treat all viewpoints equally, overlooking the fact that these camera perspectives offer different insights into the environment—some unique, some overlapping, and some redundant. Consequently, the ego vehicle may not fully capitalize on the diverse perspectives available, leading to indiscriminate collaboration that generates excessive communication and computation. Actually, communication may not be necessary when some partners share very similar observations.

Therefore, we believe that leveraging the poses can effectively guide collaboration. When a vehicle is provided with its partners' poses and their cameras' poses, it can estimate and assign varying degrees of significance to different camera viewpoints based on their spatial relationship and relevance to the ego's perception. In doing so, the ego vehicle can make informed and proactive selections about which sensory inputs to prioritize, thereby enhancing its perception capacity while reducing communication redundancy. This active selection mechanism recognizes the importance of the ego's self-awareness and adaptability and empowers ego vehicles to make contextually relevant decisions. This not only improves perception but also fosters a more dynamic, scalable, and efficient collaborative environment for autonomous robots. We will elaborate on the details in the following sections.

\subsection{Definition}

Consider $N$ agents in total each with $M$ cameras in the scene. Let $\mathcal{B}$ be the ego agent's BEV feature encoding. Let $\{\mathcal{X}_{i}\}_{i=1}^M$ and $\mathcal{Y}$ be the observation sets of all cameras and the perception supervision of ego agent, respectively. The objective of collaborative perception is to achieve the maximized perception performance of all agents as a function of the number of selected agents $N$; that is,
\begin{equation}
    \xi_{\Phi}(N)=\underset{\theta, \mathcal{P}}{\arg \max } \sum_{k=1}^N g\left(\Phi_\theta\left(\mathcal{B}\left(\{\left\{\mathcal{X}_{i}\right\}_{i=1}^{ M}\}_{k=1}^N\right)\right), \mathcal{Y}\right)
    \label{eq:collab}
\end{equation}
where $g(\cdot, \cdot)$ is the perception task, specifically it can be 3D object detection evaluation metrics such as bounding boxes for mAP evaluation. $\Phi$ is the perception network with trainable parameter $\theta$, and $\{\left\{\mathcal{X}_{i}\right\}_{i=1}^{ M}\}_{j=1}^N$ are the messages transmitted from the $j$th agent (each with $M$ features) to the ego agent. The network learns BEV representation from the camera observations via BEV feature encoding which is detailed below and outputs object detection results. Note that when $N=1$, there is no collaboration and $\xi_{\Phi}(1)$ reflects the single-agent perception performance.

\subsection{Details}
\textbf{BEV feature encoding.}
We adopt the BEV encoder backbone in BEVFormer~\cite{li2022bevformer} to obtain BEV representations and perform the 3D object detection task. BEVFormer has demonstrated state-of-the-art performance in single-car 3D object detection tasks. This architecture is based on a Transformer-based design that incorporates efficient deformable attention layers as an alternative to traditional multi-head attention mechanisms. This choice mitigates the computational cost associated with modeling extensive sequences, such as pixels or voxels. In the deformable attention layer, 3D BEV queries are transformed into 2D reference points and projected onto the respective camera's features. Therefore, it is notable that when applied to multi-agent collaboration, communication overhead scales with the number of 3D-to-2D queries, increasing in tandem with the number of partner vehicles and cameras involved.
To achieve efficient collaboration, we augment deformable attention module with a coordinate-based active selection network to assign each query with an interest score $\mathcal{I}=\{\mathcal{I}_{ki}\}_{k=1,i=1}^{N_{\text{car}},N_{\text{view}}}$ where $\mathcal{I}_{ki}\in[0,1]$. Here $N_\text {car}$ and $N_\text {view}$ mean the total number of agents and image views for each agent. We use Attn to denote each query point's attention weights result following Eq.~\ref{eq: dfm}:
\begin{equation}
    \operatorname{Attn }\left(q, F_{ki}\right)=\operatorname{DfmAttn}\left(q, \mathcal{P}_k(q, i, j), F_{ki}\right)
\end{equation}
And based on that we develop the pose-guided selective attention (PSA) layer:
\begin{equation}
    \text{PSA}\left(q, F\right)=\sum_{k=1}^{N_\text {car}} \frac{1}{\left|\mathcal{R}_{\text {hit }}\right|} \sum_{i \in \mathcal{R}_{\text {hit }}} \mathcal{I}_{ki} \sum_{j=1}^{N_{\text {ref }}} \operatorname{Attn }\left(q, F_{ki}\right)
    \label{eq:psa}
\end{equation}
Here $q\in \mathbb{R}^{1\times C}$ directly follows Eq.~\ref{eq:sca}. Note the difference is that, $\mathcal{P}_k(q, i, j)$ is specifying projection matrix of the $i$-th camera of $k$-th vehicle denoted as $\mathbf{T^{\prime}_k}\cdot \mathbf{T_i}$ with image feature $F_{ki}$ , indicating car-to-car then ego-to-image transform. $\mathcal{I}_{ki}=0$ means that query $q$ will not attend to this camera image. We introduce how $\mathcal{I}$ is obtained in the below.

\textbf{Active selection.}
As mentioned above, the active selection mechanism outputs the interest score $\mathcal{I}_{ki}\in[0,1]$ via a simple network. As illustrated in Fig.~\ref{fig:main}, it takes as input the BEV query  $q$, along with the aforementioned transformation matrices $ \mathbf{T_{c_{ki}}} = \mathbf{T^{\prime}_k}\cdot \mathbf{T_i}$. The network that outputs interest score map $\mathcal{I}$ is formulated as:

\begin{equation}
    \mathcal{I} = \sigma\left(\text{MLP}(\text{concat}(q, \text{PE}(\mathbf{T_{c_{ki}}}))\right)
\end{equation}
Here $\text{PE}(\cdot)$ embeds transform matrix to a pose embedding, and $\sigma$ represents sigmoid non-linearity which serves as a gated selection module.
The goal is to generate a collection of interest scores for each query $q$ with regards to every feature $F_{ki}$ of the $i$-th camera from $k$-th car then use it to guide collaboration. From another perspective, this also forms a BEV map $\mathcal{M}_{ki}\in\mathbb{R}^{H\times W}$ for each feature indicating what queries are interested in it, as visualized in Fig.~\ref{vis}~(A) in the experiment section.
And naturally for query $q$ at $(x,y)$: $I_{ki} = \mathcal{M}_{ki}(x,y)$.
This map serves to highlight the relevance and significance of each feature for the ego's perception.
The sigmoid gating ensures the resulting interest scores will be lining towards 0 or 1. During inference, for those queries with low interest scores, we set a threshold $\epsilon\ll1$ and only select those that have the value above it. More formally,
the query should collaborate with the $i$-th image of $k$-th vehicle only if $\mathcal{I}_{ki}>\epsilon$.

\textbf{Task specific decoder head.}
Under collaborative conditions, we generate collaborative BEV encoding with PSA. Subsequently, the BEV encoding is fed into a Deformable DETR head \cite{zhu2020deformable} for 3D bounding boxes prediction.

Further experience shows that the collaborative BEV encoding significantly improves 3D detection performance on the basis of collaboration, and effectively leverages information from different perspectives.

\section{Experiments}
\subsection{Experiment setup}
\textbf{Dataset.}
We conducted experiments using the V2X-Sim Dataset~\cite{li2022v2x}, an extensive dataset that simulates complex urban driving scenarios using the CARLA simulator~\cite{dosovitskiy2017carla}. Our training dataset comprises 80 scenes, while the validation and testing dataset consist of 10 scenes each. The dataset is sampled at a rate of 5 Hz. Furthermore, we adapted the V2X-Sim Dataset to match the data format of the nuScenes standard~\cite{caesar2020nuscenes} in the MMDet3D framework~\cite{mmdet3d2020}. We pre-processed the voxel grids within a range of $\left[-51.2m, 51.2m\right]$ in the x and y-axes and lifted $N_{\text{ref}}=4$ points on the z-axis.
We also aimed to test our method on real-world datasets. However, current options have some limitations. The DAIR-V2X dataset~\cite{yu2022dair} focuses on vehicle-to-infra collaboration, which is not applicable to our case. Additionally, the vehicle-to-vehicle dataset V2V4Real~\cite{xu2023v2v4real} has not yet released the camera data. We look forward to testing our method on real-world data once it becomes available in the future.

\textbf{Baseline.}
We conducted a direct comparison between ActFormer and the BEVFormer version 1 model, which was extended to collaboration using all queries. We refer to this baseline as Co-BEVFormer. This comparison allows us to assess the impact of the proposed active selection mechanism. The results of this comparison under different numbers of cars are presented in Table
\ref{table:performance}.
We also compare the performance with some existing collaborative perception models. They rely on different collaboration strategies and are all based on LiDAR inputs. All methods are listed in Table \ref{table:cop} for a comprehensive comparison. It is important to note that our study represents a novel approach for efficient camera-only collaborative perception, which distinguishes it from previous efficiency methods based on other modalities.

\begin{table}[t]
    \vspace{5mm}
    \caption{Collaborative 3D object detection results with different numbers of vehicles using nuScenes center distance metric.}
    \vspace{-5mm}
    \begin{center}
        \resizebox{\columnwidth}{!}{
            \begin{tabular}{c|c|c|c|c|c|c}
                \Xhline{4\arrayrulewidth}
                \multirow{2}{*}{\textbf{Paradigm}} & \multirow{2}{*}{\textbf{Method}} & \multicolumn{5}{c}{\textbf{AP with different $N_{\text{car}}$}}                                                                         \\
                \cline{3-7}
                                                   &                                  & $1$                                                             & $2$             & $3$             & $4$             & $5$             \\
                \hline
                \multirow{1}{*}{Single-agent}      & BEVFormer~\cite{li2022bevformer} & 52.1                                                            & N/A             & N/A             & N/A             & N/A             \\
                \hline
                \multirow{2}{*}{Multi-agent}       & Co-BEVFormer                     & $\textbf{52.1}$                                                 & 53.8            & 55.0            & 56.1            & 60.8            \\
                                                   & ActFormer                        & 51.7                                                            & $\textbf{54.8}$ & $\textbf{55.8}$ & $\textbf{58.9}$ & $\textbf{61.2}$ \\
                \Xhline{4\arrayrulewidth}
            \end{tabular}}
    \end{center}
    \label{table:performance nuscenes}
    \vspace{-3mm}
\end{table}

\begin{table}[t]
    \caption{Collaborative 3D object detection results for different numbers of vehicles evaluated with bounding box AP@IoU.
    }
    \vspace{-5mm}
    \begin{center}
        \resizebox{\columnwidth}{!}{

            \begin{tabular}{c|c|c|c|c|c|c}
                \Xhline{4\arrayrulewidth} \multirow{2}{*}{\textbf{Method}} & \multirow{2}{*}{\textbf{mAP}} & \multicolumn{5}{c}{\textbf{AP@IoU=0.5 / AP@IoU=0.7 with $N_{\text{car}}$}}                                                                     \\
                \cline{3-7}
                                                                           &                               & $1$                                                                        & $2$            & $3$            & $4$            & $5$            \\

                \hline
                \multirow{2}{*}{Co-BEVFormer}                              & @0.5                          & \textbf{32.54}                                                             & 34.15          & 38.74          & 43.34          & 45.88          \\
                                                                           & @0.7                          & \textbf{19.76 }                                                            & 22.16          & 24.03          & 26.05          & 29.89          \\
                \cline{1-7}

                \multirow{2}{*}{ActFormer}                                 & @0.5                          & 31.40                                                                      & \textbf{37.42} & \textbf{40.41} & \textbf{44.70} & \textbf{53.33} \\
                                                                           & @0.7                          & 19.71                                                                      & \textbf{25.48} & \textbf{31.23} & \textbf{36.31} & \textbf{45.15} \\
                \Xhline{4\arrayrulewidth}
            \end{tabular}}
    \end{center}
    \label{table:performance}
    \vspace{-5mm}
\end{table}

\textbf{Implementation details.}
Our implementation consists of a 3-layer Pose-Guided Selection Attention (PSA) network and a 3-layer temporal self-attention mechanism from the backbone of BEVFormer, serving as the encoder for our framework. For Pose Embedding (PE), we embed the 4x4 homogeneous matrix into a 256-dimensional vector, matching the feature embedding dimension.
Each query feature is concatenated with the Pose Embedding of the corresponding image. Then the result of concatenation is fed to the active selection network that consists of two linear layers to produce an interest score for each query.
During inference, we remove queries with interest scores less than a predefined threshold, chosen as $\epsilon = 0.01$. This selection operation effectively reduces the number of queries communicated among the vehicles, resulting in approximately a 40\% reduction in computations and operations.

\textbf{Evaluation metrics.}
For the 3D object detection task, we report the Average Precision (AP) under two different bounding box Intersections Over Union (IOU) thresholds, namely 0.5 and 0.7, following the settings of previous works.
Additionally, an alternative evaluation metric is utilized in the nuScenes dataset, which uses center distance difference as the mAP threshold. As BEVFormer was originally tested under this setting, we also evaluated our method using the same metric for a fair comparison. Results are listed in table \ref{table:performance nuscenes}. Our performances under the two types of metrics appear to be consistent.

\subsection{Quantitative results}
\textbf{Comparison with BEVFormer.}
We employed the BEVFormer as the single-agent detection model, without any collaboration, and refer to it as the \textit{single-agent} BEVFormer. This model achieved an mAP of 52.1 under the nuScenes metric, which is reasonably competitive compared to the results reported in the original paper.
For multi-vehicle, we trained the same backbone with inputs from all participating vehicles' images combined, which we term as \textit{Co-BEVFormer}. The detection results for both models are presented in Tables \ref{table:performance nuscenes} and \ref{table:performance} with different metrics. It is evident that ActFormer demonstrates performance improvements through collaboration and outperforms Co-BEVFormer, which simply aggregates multi-agent observations as input. The performance margin becomes even more pronounced as the number of vehicles increases, particularly under the challenging AP@IoU=0.7 metric.

\begin{figure*}[h]
    \centering
    \includegraphics[width=\linewidth]{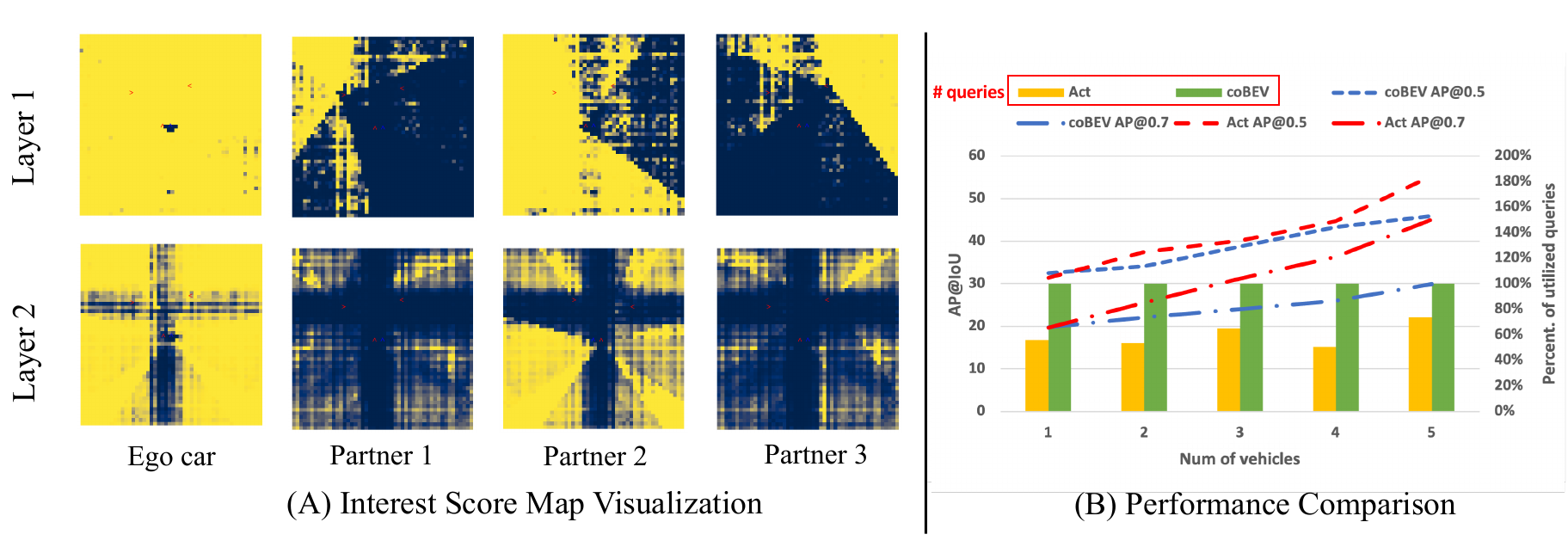}
    \vspace{-10mm}
    \caption{(A) visualizations of the interest score map of ego vehicle and its 3 partners for 2 layers of PSA. (B) a comparison of the percentage of queries used versus the performance gain under AP@IoU evaluation. Act stands for ActFormer and coBEV stands for the baseline Co-BEVFormer.}
    \label{vis}
\end{figure*}

\textbf{Comparison with the state of the arts.}
Further, we conducted a comprehensive comparison of our camera-based approach with several strong LiDAR-based methods, including When2Com, Who2Com, DiscoNet, where2comm, and V2VNet, as presented in Table \ref{table:cop}. Despite the inherent challenges of performing 3D object detection from 2D images compared to LiDAR-based detection, ActFormer surpasses two LiDAR-based baselines and outperforms the camera-based baseline. This achievement can be attributed to the development of a spatial-information-guided interest score map for queries, facilitating the exchange of only pertinent messages and thereby ensuring efficient and effective collaboration. It is also important to note that this BEV query-based active collaboration approach sets our work apart from previous efforts in the field. Table \ref{table:components} provides an illustrative comparison of the different message fusion methods employed in these collaboration systems.

\begin{table}[t]
    \vspace{5mm}
    \caption{ActFormer compared with other collaboration methods on V2X-Sim for the detection task. “L” and “C” indicate LiDAR and Camera, respectively.}
    \vspace{-5mm}
    \begin{center}
        \resizebox{\columnwidth}{!}{

            \begin{tabular}{c|c|c|c}

                \Xhline{4\arrayrulewidth}
                \multirow{2}{*}{\textbf{Method} }                   & \multirow{2}{*}{\textbf{Modality}} & \multicolumn{2}{c}{\textbf{Detection}}                       \\
                \cline{3-4}
                                                                    &                                    & \textbf{AP@IoU=0.5}                    & \textbf{AP@IoU=0.7} \\

                \hline
                \multirow{1}{*}{When2com~\cite{liu2020when2com}}    & L                                  & 44.02                                  & 39.89               \\

                \multirow{1}{*}{Who2com~\cite{liu2020who2com}}      & L                                  & 44.02                                  & 39.89               \\

                \multirow{1}{*}{Where2comm~\cite{hu2022where2comm}} & L                                  & 59.10                                  & 52.20               \\
                \multirow{1}{*}{V2VNet~\cite{wang2020v2vnet}}       & L                                  & 68.35                                  & 62.83               \\
                \multirow{1}{*}{DiscoNet~\cite{li2021learning}}     & L                                  & \textbf{69.03}                         & \textbf{63.44}      \\
                \hline
                \multirow{1}{*}{Co-BEVFormer}                       & C                                  & 45.88                                  & 29.89               \\
                \multirow{1}{*}{\textbf{ActFormer}}                 & C                                  & $\textbf{53.33}$                       & $\textbf{45.15}$    \\
                \Xhline{4\arrayrulewidth}
            \end{tabular}}
    \end{center}
    \label{table:cop}
\end{table}

\begin{table}[t]
    \caption{Comparisons of communication approaches in collaborative perception systems, following the demonstration in \cite{hu2022where2comm}.}
    \vspace{-5mm}
    \begin{center}
        \resizebox{\columnwidth}{!}{

            \begin{tabular}{c|c|c}

                \Xhline{4\arrayrulewidth} \multirow{1}{*}{\textbf{Method}} & \textbf{Message}   & \textbf{Message fusion}   \\
                \hline
                When2com~\cite{liu2020when2com}                            & Full feature map   & Attention per-agent       \\
                V2VNet~\cite{wang2020v2vnet}                               & Full feature map   & Average per-agent         \\
                DiscoNet~\cite{li2021learning}                             & Full feature map   & MLP attn per-location     \\
                Where2comm~\cite{hu2022where2comm}                         & Sparse feature map &
                Attention per-location                                                                                      \\
                \hline
                ActFormer                                                  & Active BEV queries & Deformable attn per-query \\
                \Xhline{4\arrayrulewidth}
            \end{tabular}}
    \end{center}
    \label{table:components}
\end{table}

\textbf{Efficiency.}

The proposed query selection method not only leads to substantial performance improvements but also significantly reduces the number of query points, by approximately 50\%, compared to the non-active utilization of all queries. This reduction not only minimizes communication overhead but also alleviates the computational burden, as the number of queries directly impacts the computation complexity in the subsequent attention mechanism. The percentage of query reduction for different numbers of agents compared to the Co-BEVFormer baseline is listed in Table~\ref{table:eff}, presented as the percentage $P_{N_{\text{car}}} = N_{\text{act}} / N_{\text{ori}}$. Here, $N_{\text{ori}}$ and $N_{\text{act}}$ indicate the average number of the original non-active queries and the average number of active queries for all agents, respectively. These results are averaged over the testing dataset, demonstrating that our method reduces query points for all agents, even for a single agent.

\subsection{Qualitative results}
We present visualization results of the interest score maps in Fig.~\ref{vis}. Figure~\ref{vis}~(A) illustrates the total interest score maps at two active attention layers for both the ego car and its three partner vehicles. The interest score maps of all cameras on the same vehicles are combined together for visualization. The color map ranging from blue to yellow represents the interest score, ranging from 0 to 1, indicating the level of interest for the active queries. Notably, in the first layer, the ego car chooses queries to attend to each partner vehicle in areas that are farther away from the ego's own observation. Specifically, the blue regions on the interest score maps for partners overlap with the ego's viewpoints, which have already been covered by the ego's selections, as shown by the almost entirely yellow interest score map on the left. This behavior aligns with the logic of collaboration: by querying observations that are more challenging to access, the ego car gathers additional information.
In the second layer, the network learns to identify crossroads, indicating that it has acquired a general understanding of the scene beyond the specific object detection task. It focuses its attention on vehicles within the intersection and objects outside the road. This observation underscores the significance of the interaction between BEV queries and the spatial information provided by each partner.

Figure~\ref{vis}~(B) provides a visual comparison of the percentages of queries utilized when different numbers of vehicles are collaborating, along with the corresponding detection performance. It is evident that as more vehicles participate, ActFormer consistently uses significantly fewer queries than the baseline and achieves a larger margin in detection performance. This clearly demonstrates the efficiency and effectiveness of ActFormer for multi-agent object detection.

\begin{table}[t]
    \caption{Comparison on efficiency. $N_{\text{ori}}$ and $N_{\text{act}}$ stand for the number of non-active queries of original approaches and that of active queries of ActFormer, respectively. }
    \centering
    \resizebox{\columnwidth}{!}{

        \begin{tabular}{c|c|c|c|c|c}

            \Xhline{4\arrayrulewidth} \multirow{2}{*}{\textbf{}} & \multicolumn{5}{c}{\textbf{Components of different $N_{\text{car}}$} }                                         \\
            \cline{2-6}
                                                                 & $1$                                                                    & $2$     & $3$     & $4$     & $5$     \\
            \hline
            $N_{\text{ori}}$                                     & $4.45$k                                                                & $15.8$k & $23.5$k & $27.3$k & $31.5$k \\

            \hline
            $N_{\text{act}}$                                     & $2.50$k                                                                & $8.46$k & $15.3$k & $13.7$k & $23.2$k \\
            \hline
            \textbf{$P_{N_{\text{car}}}$}                        & 55.98\%                                                                & 53.28\% & 65.01\% & 50.53\% & 73.60\% \\
            \Xhline{4\arrayrulewidth}
        \end{tabular}}

    \label{table:eff}
\end{table}

\section{Conclusion}
This paper proposes ActFormer, an efficient and scalable method for multi-robot collaborative 3D object detection from 2D images with active 3D-to-2D queries. It utilizes the poses of collaborating partners and actively selects a sparse set of BEV queries to interact with the 2D image features for BEV representation learning. Comprehensive experiments prove that it significantly reduces information redundancy while still enhancing the detection performance. We believe ActFormer is a versatile method for multi-agent perception and plan to extend it to multi-modality input and different perception tasks in our future works.


    {
        \small
        \bibliographystyle{IEEEtran}
        \balance
        \bibliography{IEEEabrv}
    }

\end{document}